%% file: bare_jrnl.tex
\documentclass[journal]{IEEEtran}
\usepackage{times}
\usepackage{epsfig}
\usepackage{graphicx}
\usepackage{amsmath}
\usepackage{amssymb}

\usepackage{cite}
\usepackage{enumitem}

\usepackage[ruled]{algorithm2e}

\usepackage{multirow}
\usepackage{makecell}
\usepackage{booktabs}
\usepackage{color}

\newcommand{\etal}{\textit{et al.}}
\newcommand{\ie}{\textit{i.e.}}
\newcommand{\eg}{\textit{e.g.}}
\definecolor{purple}{rgb}{0.54, 0.17, 0.89}
\ifCLASSINFOpdf
\else
\fi
\begin{document}
%
\title{Adversarial Attacks on LiDAR-Based Tracking Across Road Users: Robustness Evaluation and Target-Aware Black-Box Method}
%
%
%

\author{Shengjing~Tian,
        Xiantong Zhao,
        Yuhao Bian,
        Yinan Han*,
        and~Bin Liu
\thanks{This work has been submitted to the IEEE for possible publication. Copyright may be transferred without notice, after which this version may no longer be accessible.}
\thanks{Shengjing~Tian is with the School of Economics and Management, China University of Mining and Technology, China. }
\thanks{Yinan Han is with the DUT-BSU Joint Institute, Dalian University of Technology, China. Corresponding author: spolico\_hyn@outlook.com}
\thanks{Bin Liu is with the School of Mathematics and Information Science, Nanchang Hangkong University, Nanchang, China.}
\thanks{Yuhao Bian and Xiantong Zhao are with the School of Mathematical Sciences, Dalian University of Technology, China.}
}


%
%

\markboth{This work has been submitted to the IEEE for possible publication. Copyright may be transferred without notice, after which this version may no longer be accessible.}%
{Shell \MakeLowercase{\textit{et al.}}: Bare Demo of IEEEtran.cls for IEEE Journals}
%



\maketitle

\begin{abstract}
In this study, we delve into the robustness of neural network-based LiDAR point cloud tracking models under adversarial attacks, a critical aspect often overlooked in favor of performance enhancement. 
These models, despite incorporating advanced architectures like Transformer or Bird's Eye View (BEV), tend to neglect robustness in the face of challenges such as adversarial attacks, domain shifts, or data corruption. 
We instead focus on the robustness of the tracking models under the threat of adversarial attacks.
We begin by establishing a unified framework for conducting adversarial attacks within the context of 3D object tracking, which allows us to thoroughly investigate both white-box and black-box attack strategies. 
For white-box attacks, we tailor specific loss functions to accommodate various tracking paradigms and extend existing methods such as FGSM, C\&W, and PGD to the point cloud domain.
In addressing black-box attack scenarios, we introduce a novel transfer-based approach, the Target-aware Perturbation Generation (TAPG) algorithm, with the dual objectives of achieving high attack performance and maintaining low perceptibility.
This method employs a heuristic strategy to enforce sparse attack constraints and utilizes random sub-vector factorization to bolster transferability.
Our experimental findings reveal a significant vulnerability in advanced tracking methods when subjected to both black-box and white-box attacks, underscoring the necessity for incorporating robustness against adversarial attacks into the design of LiDAR point cloud tracking models. Notably, compared to existing methods, the TAPG also strikes an optimal balance between the effectiveness of the attack and the concealment of the perturbations.
\end{abstract}


%
\IEEEpeerreviewmaketitle

\input{introduction.tex}
\input{related_work.tex}
\input{proposed_method.tex}

\input{experiments.tex}
\input{conclution.tex}

\section*{Acknowledgment}
This work is supported in part by the National Natural Science Foundation of China (Grand No. 62301562, 61976040), the China Postdoctoral Science Foundation (Grand No. 2023M733756), the Fundamental Research Funds for the Central Universities (Grand No. 2023QN1055), and the China Scholarship Council. We thank the reviewers for their valuable comments toward improving this paper.

\ifCLASSOPTIONcaptionsoff
  \newpage
\fi



\bibliographystyle{IEEEtran}
\bibliography{mybib}

\vfill


\end{document}

%% file: introduction.tex
\section{Introduction}
Point clouds acquired by Light Detection and Ranging (LiDAR) have become important intermediaries, including indoor \cite{guo_lidarnet_2024} and outdoor perception \cite{Argoverse2019, nuscenes}. In this context, visual object tracking in LiDAR point clouds is one of the critical tasks in 3D vision for understanding the environment, which utilizes 3D bounding boxes to locate objects of interest continuously in successive frames. In recent years, deep neural networks (DNN) have enabled significant advances in accuracy and precision for point cloud tracking tasks. Although they have succeeded tremendously, 3D trackers based on DNNs lack comprehensive robustness studies. Nowadays, many investigations \cite{Liu2022Graph, chen_universal_2020, li_fooling_2021, cao_adversarial_2019, tu_physically_2020, hu_pointca_2022, sun_towards_2020} reveal that current DNNs are still vulnerable to adversarial attacks, which deceive the trained model to produce incorrect results using adversarial examples injected by imperceptible noise. This also remains a major concern for point cloud tracking, especially deployed in safety-critical applications such as autonomous driving, robotics, and railway intrusion detection.
\begin{figure}[t]
    \centering
    \includegraphics[width=1\linewidth]{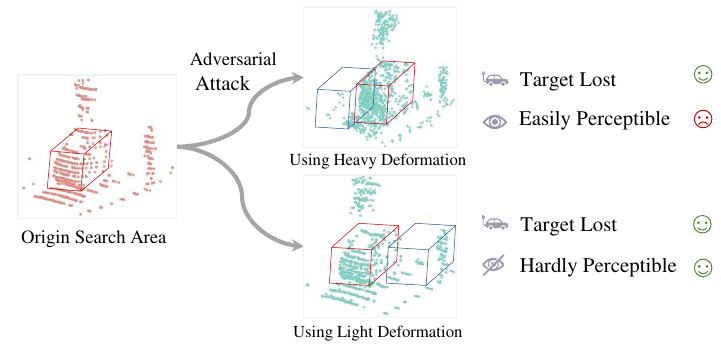}\\
    (a) The goal of the adversarial attack\\
    \includegraphics[width=1\linewidth]{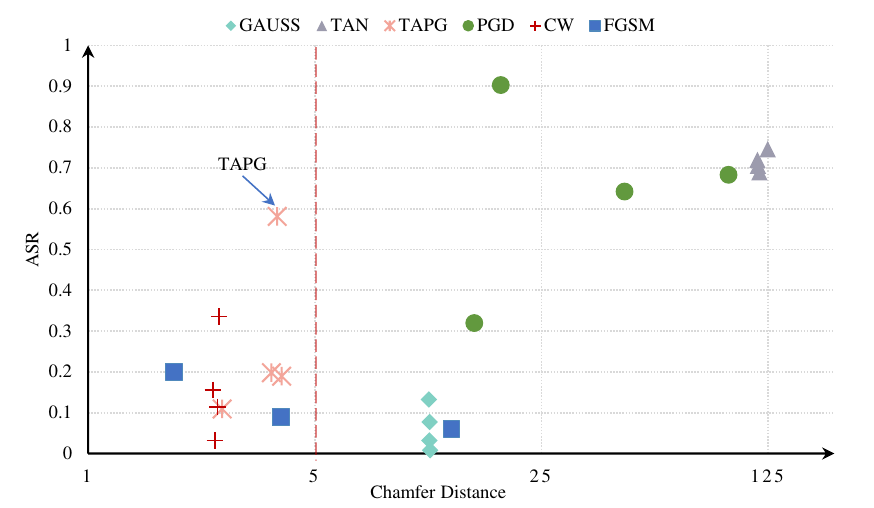}\\
    (b) The performance of different attack methods.
    \caption{The goals of the adversarial attack. (a) demonstrates the motivation of this work, \ie, the samples generated by the adversarial attack need to make the original model lose its target and maintain its own imperceptibility. (b) gives the results of the proposed TAPG compared to the other attack methods in terms of the attack success rate (ASR) and Chamfer distance (CD). The closer the method is to the top-left corner, the more superior it is. }
    \label{fig:bubble_all}
\end{figure}

Adversarial attacks have received much attention due to their important role in enhancing model robustness \cite{CW}. Generally, adversarial attacks can be classified into white-box and black-box attacks. The former requires explicit information about the network structure, pre-training weights, and even loss function. The latter has no access to any information other than model inputs and outputs. Currently, in images and point clouds, many white-box attacks have been investigated deeply for classification tasks, giving rise to some representative attack frameworks such as FGSM \cite{FGSM}, PGD\cite{PGD}, and C\&W\cite{CW}. In real-world scenarios, however, the victim models are often only queried, and hardly obtain their detailed information.

In light of this, some researchers have shifted the attention to black-box attacks\cite{Shi2022QueryEfficientBA, liu_multiview_2021, Ding2024TransferableAA, jia_iou_2021}. Compared with white-box attacks, it cannot calculate the gradients that could guide the generation of perturbations, thus sparking out three frameworks: gradient estimation-based methods, decision boundary-based methods, and transfer-based methods. Gradient estimation-based methods, called zeroth order optimization methods, approximate the gradient with the numerical difference between samples under small perturbations. Decision boundary-based methods take adversarial samples with heavy noise as the origin, compress the corresponding noise, and query the model after each compression to determine whether they fall within the discriminant boundary. The above two methods are usually adopted in the image domain, especially in classification tasks\cite{Shi2022QueryEfficientBA}. To cope with complex perception tasks in different modalities, the transfer-based approaches assume the existence of a model that knows its details and then apply the samples generated by this surrogate model via a white-box attack into the target model. In this work, DNN-based visual tracking from point clouds not only suffers from the irregular structure, but also complicated network architecture comprising feature extraction, correlation component, and location head. Therefore, we focused on devising the transfer-based attack because of its flexibility for complex perception tasks. 

Recently, many point cloud tracking methods have been proposed to exert the advantages of deep learning, such as P2B \cite{P2B}, M2Track\cite{MMTrack}, V2B \cite{V2B}, and MBPTrack\cite{MBPTrack}. These methods extract geometric and appearance features by PointNet++\cite{PointNetPP2017} or DGCNN \cite{DGCNN}, then leverage scale-aware information, surface completion, or memory mechanism to generate bounding boxes for 3D targets. Even though they achieve impressive accuracy performances, the robustness against adversarial samples is untapped comprehensively for transfer-based attacks.  
There are two main challenges: how to guarantee the imperceptible ability of adversarial samples derived from the substitute model, and how to improve the transfer ability of adversarial samples when applied to the target model. Please refer to Fig. \ref{fig:bubble_all}.

To address the above problems, we propose a black-box attack method for the advanced trackers in point clouds, which can generate adversarial samples with target-aware perturbation factorization to interfere with the norm results while maintaining the concealment of the adversarial noises. This intuition takes inspiration from sparse vector factorization in the classification tasks of images\cite{Fan2020SparseAA}. Specifically, the DNN-based 3D trackers construct different implicit functions for the target location prediction, and imperceptible adversarial perturbation from the surrogate tracker may be hardly suitable for the target tracker. Therefore, we improve the transfer ability of the originally adversarial perturbation by constraining its decomposed sub-vectors. Theoretically, the original perturbation is generally closer to the decision boundary of its host, and certain offsets of the decision boundary exist among different trackers. To cross this offset, we push the decomposed two sub-vectors to the vicinity of the decision boundary by perception loss and adversarial loss, enabling a stronger transfer ability for the origin perturbation. In addition, we compare the transfer ability of the adversarial samples generated by the surrogate model when it employs the frameworks of the FGSM, PGD, and CW. To do this, we must design the appropriate loss and reproduce the three attacks under the pipeline of the point cloud. Finally, we utilize four advanced tracking models (P2B, BAT, STNet, and M2Track) to evaluate their robustness in large-scale outdoor scenarios including KITTI \cite{KITTI2013} and nuScenes \cite{nuscenes}. Our experimental results, on the one hand, demonstrate the proposed transfer-based method can generate high-quality adversarial samples to achieve an efficient attack, on the other hand, we draw a conclusion that the current 3D trackers indeed suffer from the vulnerability in digital attack and adversarial training is urgent to be considered.

To sum up, the contributions of this work are as follows:
\begin{itemize}
    \item We are the pioneer in building a benchmark that evaluates the robustness of DNN-based 3D trackers under both white-box and black-box attack frameworks.
    \item We innovatively propose a black-box attack algorithm for DNN-based 3D trackers, which leverages target-aware perturbation factorization to improve the transferability of adversarial samples.
    \item We integrate the proposed method into three representative 3D trackers, whose adversarial samples possess stronger concealment and transferability than the existing black-box attack method.
\end{itemize}

%% file: related_work.tex
\section{Related Work}
\subsection{Visual Tracking in Point Clouds}
Visual tracking in point clouds benefits from the property of target tracking for arbitrarily specified templates, which has been widely studied. Early work \cite{itsc} is based on the mean-shift approach to estimate the 3D bounding box of a target over two modalities, point cloud and image. Pang \etal \cite{Modelfreetrack} proposed a vehicle tracking method based on an optimization solution that adopts geometric constraints using physical prior.

However, the above traditional methods are not satisfactory in terms of tracking accuracy to complex scenarios. The development of deep learning methods has injected new vigor into 3D vision. Under the framework of the Siamese networks, current trackers can now be mainly summarised as shape-appearance feature matching\cite{Siam3D2019}, motion guidance \cite{MMTrack}, information embedding \cite{P2B, BAT}, and bird's-eye view (BEV) transformation\cite{V2B, stnet}. 
Initially, due to the unstructured representation nature of point cloud data, some researchers obtained the BEV of the point cloud and then applied the 2D convolution operation on it\cite{FastAF}. This method may lose the geometric information. Afterward, V2B\cite{V2B} and STNet\cite{stnet} propose to utilize BEV on high-level semantic features. 

On the other hand, point-based DNNs provide new representation forms to deal with unstructured point cloud data. Specifically, considering the success of the Siamese network in the 2D tracking, Giancola \etal \cite{Siam3D2019} resorted to PointNet \cite{PointNet2017} to learn a generic matching function for the appearance of the target. However, it relies on probabilistic sampling for the generation of candidate point clouds, whose training process is slow and expensive. To address this problem, both P2B\cite{P2B} and BAT\cite{BAT} replace the traditional sampling with deep Hough vote \cite{DeepHough2019} to directly generate candidate bounding boxes. Subsequently, researchers improved the P2B\cite{P2B} or BAT\cite{BAT} by combining technical approaches such as sparse convolution \cite{LTTR}, Transformers \cite{PTT2021, PTTR}, and memory mechanism \cite{MBPTrack}. Although achieving considerable performance in success and precision metrics, these types of DNN-based approaches pay attention to normal scenarios, ignoring their adversarial robustness. In this work, we complete robustness evaluation experiments for 3D LiDAR trackers, which investigate the performance of current methods under white-box and black-box attacks.

\subsection{Adversarial Attacks on Image Domain}
Image classification models are initially used to discover the susceptibility to adversarial attacks, \ie, by adding some imperceptible perturbations for human beings to make the output of the model wrong or uncertain. Szegedy \etal \cite{first_attack} first found the counter-intuitive property that deep neural networks usually learn a discontinuous mapping between inputs and outputs, which can lead to misclassifying an image under subtle perturbation. Next year, FGSM \cite{FGSM}, which utilizes the gradient direction to implement adversarial attacks, is proposed to attack image classification models. Subsequently, PGD \cite{PGD} adopts multi-step updates to conduct projected gradient descent on adversarial loss, and MI-FGSM \cite{MIFGSM} stabilizes update directions by a momentum-based iterative scheme. Differently, C\&W \cite{CW} formulates the attack task as optimization problems under different constraints that can be solved by existing optimization algorithms. 
In addition to the above whit-box attack, some black-box attack technologies are also innovating in the image domain. AoA \cite{chen_universal_2020} finds that the universality of the attention mechanism can be used to conduct transfer-based attacks for multiple victim DNNs. It maximizes the difference in the attention region between benign and adversarial images. Shi \etal \cite{Shi2022QueryEfficientBA} proposed a customized iteration and sampling strategy to probe the decision boundary within limited queries. Furthermore, some work leverages transformations \cite{jiang_physical_2023} and Perturbation Factorization \cite{Fan2020SparseAA} to generate transferable adversarial samples.

For safety-critical applications, there also exist some scene understanding works to investigate the robustness of DNNs, such as object detection \cite{Li2018RobustAP, liu_multiview_2021, Ding2024TransferableAA}, person re-identification \cite{bai_adversarial_2020}, semantic segmentation \cite{xie2017adversarial}, and 2D visual tracking \cite{yan_cooling_2020, liang_efficient_2020, jia_robust_2020, guo_spark_2020, jia_iou_2021}. Herein, we briefly review the 2D visual tracking as it has a similar purpose to our work, but operates on different modalities. Guo \etal \cite{guo_spark_2020} proposed a spatial-aware online incremental attack by minimizing the overlap between the target box and the adversarial box. Jia \etal \cite{jia_robust_2020} minimized the confidence difference between the correct and incorrect binary samples and initialized the perturbation of the current frame with the one generated by the previous frame. Yan \etal \cite{yan_cooling_2020} trained a perturbation generator using a cooling-shrinking loss. Jia \etal \cite{jia_iou_2021} proposed a black-box attack method for visual tracking, which starts from a heavy noisy image and gradually compresses the noise level from both tangential and normal directions. 

\subsection{Adversarial Attacks on Point Clouds}
Point cloud data is usually collected by LiDARs or depth cameras, which have become a critical modality to percept the surroundings. Even though the above methods in the image domain have made great progress, the irregular structure of point clouds has rendered 2D image attack algorithms ineffective. Firstly, in raw point clouds, there is no grid-like arrangement of images that can be subtly altered; Secondly, the perturbation vectors lie in the vast search space since they can appear at any 3D location; Thirdly, inconsistency and non-correspondence in the number of target points between two frames does not exist in the image domain.

There is much adversarial attack research for DNN-based point cloud analysis, such as point cloud classification \cite{xiang_generating_2019, liu_adversarial_2020, kim_minimal_2021}, completion\cite{hu_pointca_2022}, and detection\cite{tu_physically_2020}. Xiang \etal \cite{xiang_generating_2019} attacked the classification model over PointNet, which adopts the C\&W framework to generate new points or perturb old points. Liu \etal \cite{liu_adversarial_2020} focused on imperceptible perturbations that are generated by morphing the shape of the point cloud. Kim \etal \cite{kim_minimal_2021} aimed at preserving the origin shape via perturbing a minimal number of points. In addition, Hamdi \etal \cite{AdvPC} and Liu \etal \cite{Liu2023Factorization} proposed to improve the transferability of adversarial samples by reconstruction and random perturbation factorization, respectively. However, for point cloud tracking, adversarial attacks require generating a misleading location rather than negative class labels of the classification task. In light of this, Liu \etal \cite{liu_transferable_2023} trained an adversarial generator using multi-fold drift loss. This method tends to produce easily perceived adversarial samples, which usually result in a larger gap relative to the benign one. In this work, we generate adversarial samples with target-aware perturbation factorization while maintaining the concealment of the adversarial noises. 

%% file: proposed_method.tex
\begin{figure}[th]
    \centering
    \includegraphics[width=1\linewidth]{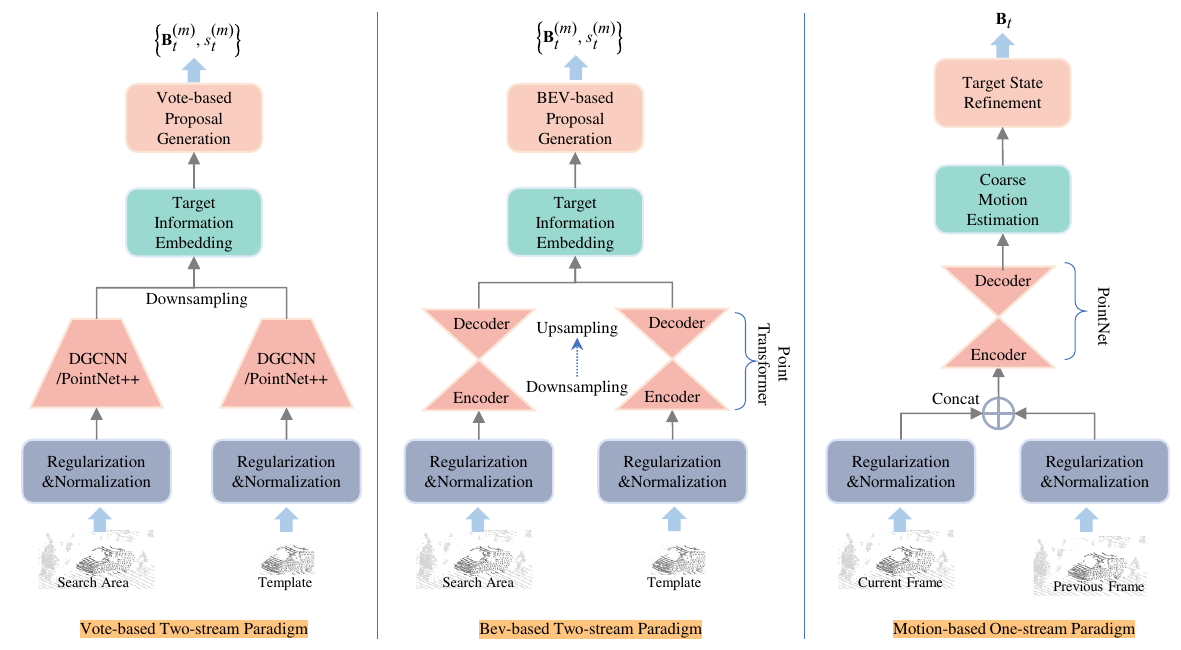}
    \caption{Different tracking paradigms for 3D LiDAR point clouds. From left to right, there are vote-based two-stream paradigm, BEV-based two-stream paradigm, and motion-based one-stream paradigm.}
    \label{fig:tracking_summary}
\end{figure}
\section{Methodology}
In this section, we first formulate the adversarial attacks of different 3D trackers into a unified framework including FGSM \cite{FGSM}, PGD\cite{PGD}, and C\&W\cite{CW}. Then, for the victim models comprising different architectures, we introduce specific loss functions that constrain the generation of adversarial samples. Last but not least, we propose a target-aware perturbation factorization to enable the transfer-based black-box attack. 

\begin{figure*}[th]
    \centering
    \includegraphics[width=0.9\linewidth]{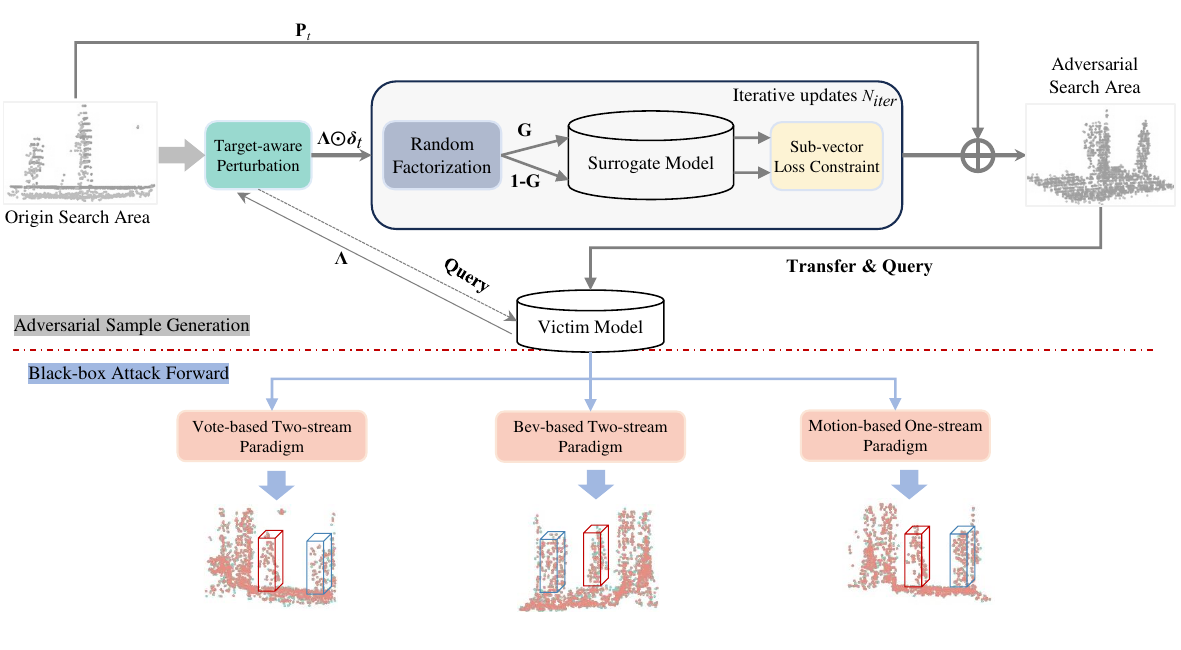}
    \caption{The pipeline of the proposed black-box attack method.}
    \label{fig:attack_pipeline}
\end{figure*}

\subsection{Problem Formulation}
In the open world, the stream of point clouds collected by the LiDAR over a certain period $T$ can be denoted as $\mathcal{S}=\{\mathbf{P}_0, \mathbf{P}_1, \dots, \mathbf{P}_t, \dots, \mathbf{P}_T\}$, where $\mathbf{P}_t \in \mathbb{R}^{N_t \times 3}$ is the $t$-th frame of point cloud video $\mathcal{S}$. For each object, we utilize a 3D bounding box $\mathbf{B}_t$ to represent it, which can be parameterized by center coordinate $(x_t, y_t, z_t)$, size $(w_t, l_t, h_t)$, and yaw angle $\alpha_t$. Given the object of interest $\mathbf{B}_0$ in the first frame, 3D tracking models $f$ aim to estimate the target state successively for each coming frame and require a high intersection-over-union (IoU) between the predicted and ground-truth bounding boxes: 
\begin{equation}\label{eq:tracker}
  IoU(\mathbf{B}_t, \mathbf{B}_t^* ) \geq \eta,
\end{equation}
where $IoU(a, b) = \frac{a \cap b}{a\cup b}$, $\eta \in [0, 1]$ is a threshold of successful tracking and $\mathbf{B}_t$ is the predicted bounding boxes. Specifically, taking a frame $\mathbf{P}_t$ as input, 3D tracking model $f$ usually produces $M$ candidates that is composed of a bounding box $\mathbf{B}_t^{(i)}$ and corresponding confidence $s^{(i)}$. And the final $\mathbf{B}_t$ is the $\mathbf{B}_t^{(i)}$ with the highest $s^{(i)}$.

The adversarial attack for 3D tracking models is to calculate a perturbation $\boldsymbol{\delta}_t$ for the $t$-th frame, making it predict an inaccurate bounding box. This progress can be formulated as follows:
\begin{equation}\label{eq:attack_origin}
  \begin{aligned}
  \min_{\boldsymbol{\delta}_t} &\quad  \mathcal{D}(\mathbf{P}_t+\boldsymbol{\delta}_t, \mathbf{P}_t), \\  
  s.t. &\quad  IoU[f(\mathbf{P}_t+\boldsymbol{\delta}_t, \mathbf{P}_0, \mathbf{B}_0), \mathbf{B}_t^*] < \eta,
  \end{aligned}
\end{equation}
where $\mathcal{D}$ is the distance metric that evaluates the error between benign and adversarial point clouds, such as Chamfer Distance \cite{CD}. Note that, if we replace the $f$ with a surrogate model $g$, this becomes a transfer-based black-box attack, \ie, using the perturbation from $g$ to attack $f$.

To conduct the white-box attack for point cloud tracking, we can intuitively extend algorithms in the image domain into this realm, such as FGSM\cite{FGSM}, PGD\cite{PGD}, and C\&W\cite{CW}. Its main problem lies in designing specific loss for different 3D trackers, which will be introduced in subsection \ref{sec:threat_models}. In this work, we focus on attacking $f$ resorting to a surrogate model $g$, which will be presented in subsection \ref{sec:target_aware}.

\subsection{White-box Attack on Victim Models}\label{sec:threat_models}
The current DNN-based trackers pay attention to the success and precision metrics. To further evaluate their robustness against adversarial samples, we choose four representative 3D trackers from the perspective of vote-based two-stream method, BEV-based two-stream method, and motion-based one-stream method. Fig.\ref{fig:tracking_summary} summary these different tracking paradigms. In the following, we review them separately and introduce the corresponding adversarial losses under white-box attacks. 

\textbf{P2B}\cite{P2B} is the first attempt that attains the goal of end-to-end training under the Siamese network. As shown in Figure[], the pipeline can be split into three modules \cite{ClassAgnostic_Tracking}: feature extraction, target information embedding, and proposal generation. Let $\{(\mathbf{B}_t^{(1)}, s_t^{(1)}), \dots, (\mathbf{B}_t^{(m)}, s_t^{(m)}), \dots, (\mathbf{B}_t^{(M)}, s_t^{(M)}) \}$ represent those candidates from the proposal generation module. To implement the white-box attack, the IoU operation involves calculating the union and interaction volumes between two cuboids, over which the gradient for backpropagation is challenging. Therefore, we design a loss function $\mathcal{H}$ with respect to the confidence $s_t^{(i)}$, and make it serve as a substitute term of $IoU[f(\mathbf{P}_t+\boldsymbol{\delta}_t, \mathbf{P}_0, \mathbf{B}_0), \mathbf{B}_t^*] < \eta$. The specific formulation can be written as: 
\begin{equation}\label{eq:attack_p2b}
  \begin{aligned}
  \mathcal{H} = -\sum_i &\big( \mathbb{I}(\| \mathbf{B}_t^{(i)} - \mathbf{B}_t^* \|_2 > \sigma) \log(s_t^{(i)}) \\
              &+ (1-\mathbb{I}(\| \mathbf{B}_t^{(i)} - \mathbf{B}_t^* \|_2 > \sigma)) \log(1-s_t^{(i)}) \big),
  \end{aligned}
\end{equation}
where $\mathbb{I}(\cdot)$ is the indicator function, $\sigma$ is a threshold of center error between two cuboids. This equation enforces the candidate whose center discrepancy with the ground-truth one is larger than $\sigma$ to predict high confidence. 

\textbf{BAT}\cite{BAT} improves the above P2B by introducing efficient geometric information. Specifically, it defines a conception of the box cloud which estimates distances from each scanned raw point to nine key points (\ie, eight vertexes and one center of a 3D bounding box).
Because its proposal generation module remains the same as the P2B, we can also utilize Equation (\ref{eq:attack_p2b}) to conduct the white-box attack. 

\textbf{STNet}\cite{stnet} is a representative method that integrates both the Transformer and BEV into the 3D tracking framework. In particular, it leverages a Siamese Point Transformer to extract the high-level features and replaces the deep Hough vote with a BEV-based localization head. Unlike the P2B and BAT, the BEV-based proposal generation module will predict a response map and take its ``pixels'' as anchors to estimate candidate bounding boxes. To attack such types of trackers, we propose shifting the peak distribution of the target response map. In addition, considering the imbalance issue of positive and negative anchors, we adopt an improved focal loss function to meet $IoU[f(\mathbf{P}_t+\boldsymbol{\delta}_t, \mathbf{P}_0, \mathbf{B}_0), \mathbf{B}_t^*] < \eta$:
\begin{equation}\label{eq:attack_stnet}
  \begin{aligned}
  \mathcal{H} &= -\sum_i \sum_j \bigg(  \mathbb{I}\big(\mathbf{M}^{(i,j)}=1 \big) \big( 1-\mathbf{r}_t^{(i,j)} \big)^\beta \log\big(\mathbf{r}_t^{(i,j)}\big) \\
                             &+\mathbb{I}\big(\mathbf{M}^{(i,j)} \neq 1\big) \big( 1-\mathbf{M}^{(i,j)} \big)^\gamma \big( \mathbf{r}_t^{(i,j)} \big)^\beta \log\big(1-\mathbf{r}_t^{(i,j)}\big) \bigg),
  \end{aligned}
\end{equation}
where $\mathbf{r}$ is the predicted response map by STNet, $\beta$ and $\gamma$ are modulation parameters of focal loss \cite{focal_loss}, and $\mathbf{M}$ is the crafted label that enables the tracker to yield adversarial samples. Its detailed formulation can be written as:
\begin{equation}\label{eq:attack_stnet_M}
\mathbf{M}^{(i,j)} = \left\{
  \begin{aligned}
  1\quad\quad\quad\quad                 &\quad i=-c_x; j=-c_y \\
  \frac{1}{\|(i, j) + (c_x, c_y)\|_2}   &\quad otherwise
  \end{aligned}
\right. ,
\end{equation}
where $(c_x, c_y)$ is the index which the target object localizes at. This formula aims to move in the opposite direction to generate peaks. 
 
\textbf{M2Track}\cite{MMTrack} points out that the appearance of the scanned object may change as LiDAR moves. The only utilization of appearance information is not sufficient in a dynamic scenario where self-occlusion easily occurs. Thus, it proposes a motion-centric paradigm, which can exert the advantage of motion clues in point cloud video. Mathematically, this method firstly learns spatiotemporal point cloud representations \cite{4DNet} to segment foreground points from $\mathbf{P}_{t-1}$ and $\mathbf{P}_{t}$. Then, the foreground points are utilized to predict the relative motion $\boldsymbol\Delta_{(t-1, t)}=(\Delta_x, \Delta_y, \Delta_z, \Delta_\alpha)$ between two consecutive frames. To conduct the white-box attack, we design the following loss $\mathcal{H}$ such that $IoU[f(\mathbf{P}_t+\boldsymbol{\delta}_t, \mathbf{P}_0, \mathbf{B}_0), \mathbf{B}_t^*] < \eta$ when $\mathcal{H}=0$:
\begin{equation}\label{eq:attack_m2track}
  \begin{aligned}
  \mathcal{H} = \| (\Delta_x, \Delta_y, \Delta_z) - (\Delta_x^* + w, \Delta_y^* + l, \Delta_z^* + h) \|_1 ,
  \end{aligned}
\end{equation}
where $(\Delta_x^*, \Delta_y^*, \Delta_z^*)$ is the ground-truth motion, $(w, l, h)$ is the size of target object. 

For different types of trackers, we can conduct white-box attack experiments by using the corresponding loss $\mathcal{H}$ under the framework of Equation \ref{eq:attack_origin}. Its results, solved by FGSM, PGD, and C\&W, will be analyzed in detail in Section \ref{experiments}. Although the white-box attack opens the way to explore the robustness problems of current methods, it requires preliminary knowledge of the network structure and pre-training weights. In addition, as claimed in \cite{feature_aware}, adversarial samples often are overfitting to origin models, leading to weak transferability. To better fit the practical application scenario, we further propose a black-box attack algorithm that is based on the philosophy of transfer learning.

\begin{algorithm}[!ht]
\caption{Target-aware Perturbation Generation}
\label{alg:solver}
\LinesNumbered
\KwIn{Point cloud sequences:$\{\mathbf{P}_t | t=0, \cdots, T \}$, Template box: $\mathbf{B}_0$}
\KwOut{Adversarial samples: $\{\Tilde{\mathbf{P}}_t \}$}
\For{$t=1, \cdots, T$}{
    $\mathbf{\boldsymbol\delta}_t \leftarrow \mathcal{N}(0, 0.05)$;\\
    \textit{// Target-aware perturbation}\\
    $\mathbf{B}_t \leftarrow \max_{s_t} f(\mathbf{P}_t, \mathbf{P}_0, \mathbf{B}_0)$;\\
    $\boldsymbol\Lambda \leftarrow \texttt{getMaskInBox}(\mathbf{P}_t, \mathbf{B}_t)\in \{0, 1\}^{N_t \times 3}$;\\
    $\Tilde{\mathbf{P}}_t \leftarrow \mathbf{P}_t + \boldsymbol\Lambda \odot \boldsymbol\delta_t$;\\
    $\Tilde{L} \leftarrow \infty$; \\
    $p \leftarrow 0.5$;\\
    $\boldsymbol\delta_t^{(0)}  \leftarrow 0$;\\
    \textit{// Iteration updates}\\
    \For{$i=1, \cdots, N_{iter}$}{
        $\Tilde{\mathbf{P}}_t^{(i)} \leftarrow \Tilde{\mathbf{P}}_t  + \boldsymbol\delta_t^{(i-1)}$;\\
        $\mathbf{G} \leftarrow \texttt{RandomSample}(\mathcal{G}, 0.5)$;\\
        \textit{// Applying substitute term of the IoU} \\
        $L \leftarrow \texttt{ChamferDist}(\Tilde{\mathbf{P}}_t^{(i)}, \mathbf{P}_t)+ \lambda\mathcal{H}\big(\Tilde{\mathbf{P}}_t^{(i)}\big)$ \\
        \qquad $+ \lambda\mathcal{H}\big(\mathbf{G} \odot \boldsymbol\delta_t^{(i-1)} + \Tilde{\mathbf{P}}_t \big) $\\
        \qquad $+ \lambda\mathcal{H}\big((1-\mathbf{G}) \odot \boldsymbol\delta_t^{(i-1)} + \Tilde{\mathbf{P}}_t \big)$;\\
        $\boldsymbol\delta_t^{(i)} \leftarrow \boldsymbol\delta_t^{(i-1)} - \xi \nabla L$;\\
        
        \eIf{$L > \Tilde{L}$}{
            $\texttt{Break}$
        }{$\Tilde{L} \leftarrow L$}
        $i\leftarrow i+1$;\\
    }
    $\Tilde{\mathbf{P}}_t \leftarrow \Tilde{\mathbf{P}}_t + \boldsymbol\delta_t^{(i)}$;\\
    $t \leftarrow t+1$    
}
*\texttt{getMaskInBox} returns 1 for points inside $\mathbf{B}_t$ and 0 for others.  \\
*\texttt{RandomSample} returns mask for factorization with a probability of 0.5. \\
*\texttt{ChamferDist} is a specific metric of $\mathcal{D}$. 
\end{algorithm}
\subsection{Black-box Attack: Target-aware Perturbation Generation}\label{sec:target_aware}
The transfer-based black-box attack approach requires a surrogate model to be selected first. In this work, P2B serves as $g$ to replace $f$ in Equation \ref{eq:attack_origin} mainly because it is the pioneer in fulfilling the end-to-end tracking and the model structure and parameters are convenient to obtain currently. In order to make its generated adversarial samples adapt to several different trackers, we propose a target-aware perturbation generation method (TAPG) empowered by a random sub-vector constraint. The whole pipeline is shown in Fig.\ref{fig:attack_pipeline}.

The goal of the TAPG contains two aspects: imperceptibility and transferability. To attain the first goal, we hope that only a few points are perturbed, \ie, sparse attack, which humans cannot precept easily. Thus, we introduce a perturbation position mask $\boldsymbol{\Lambda}$. This will change the Equation \ref{eq:attack_origin} to the following formula:
\begin{equation}\label{eq:attack_factorization_1}
  \begin{aligned}
  \min_{\boldsymbol{\boldsymbol{\Lambda}, \delta}_t} &\quad  \mathcal{D}(\mathbf{P}_t+\boldsymbol{\Lambda} \odot \boldsymbol{\delta}_t, \mathbf{P}_t) + \lambda \|\boldsymbol{\Lambda}\|_0,  \\
  s.t. &\quad  IoU[f(\mathbf{P}_t+\boldsymbol{\Lambda} \odot \boldsymbol{\delta}_t, \mathbf{P}_0, \mathbf{B}_0), \mathbf{B}_t^*] < \eta, \\
       &\quad  \boldsymbol{\Lambda} \in \{0, 1\}^{N_t \times 3},
  \end{aligned}
\end{equation}
where $\odot$ means the element-wise product, and $\lambda$ is a balance factor. 

From the perspective of transferability, the initial perturbation typically resides nearer to the decision boundary of its host tracker, and discrepancies in these boundaries are observed across various trackers. To overcome this discrepancy, we guide the two separated sub-vectors toward the proximity of the decision boundary using adversarial losses. Given a mask for factorization $\mathbf{G}$, its elements will be set to 1 with the probability $p$ and 0 with $1-p$. Let $\mathcal{G}$ be the set of all possible $\mathbf{G}$. We require the IoU for both sub-vectors factorized by the $\mathbf{G}$ to be lower than the threshold. This can be expressed as solving the expectation for all masks: 
\begin{equation}\label{eq:attack_E}
  \begin{aligned}
  &\mathop{\mathbb{E}}\limits_{\mathbf{G}\in\mathcal{G}} \big(IoU[f(\mathbf{P}_t+\mathbf{G} \odot \boldsymbol{\Lambda} \odot \boldsymbol{\delta}_t, \mathbf{P}_0, \mathbf{B}_0), \mathbf{B}_t^*] \big) < \eta  \\
  &\mathop{\mathbb{E}}\limits_{\mathbf{G}\in\mathcal{G}} \big(IoU[f(\mathbf{P}_t+(1-\mathbf{G}) \odot\boldsymbol{\Lambda} \odot \boldsymbol{\delta}_t, \mathbf{P}_0, \mathbf{B}_0), \mathbf{B}_t^*] \big) < \eta
  \end{aligned}
\end{equation}

The above Equation (\ref{eq:attack_E}) can be integrated into Equation (\ref{eq:attack_factorization_1}) as two constrain terms:
\begin{equation}\label{eq:attack_G}
  \begin{aligned}
  \min_{\boldsymbol{\boldsymbol{\Lambda}, \delta}_t} &\quad  \mathcal{D}(\mathbf{P}_t+\boldsymbol{\Lambda} \odot \boldsymbol{\delta}_t, \mathbf{P}_t) + \lambda \|\boldsymbol{\Lambda}\|_0,  \\
  s.t. &\quad  \mathop{\mathbb{E}}\limits_{\mathbf{G}\in\mathcal{G}} \big(IoU[f(\mathbf{P}_t+\mathbf{G} \odot \boldsymbol{\Lambda} \odot \boldsymbol{\delta}_t, \mathbf{P}_0, \mathbf{B}_0), \mathbf{B}_t^*] \big)< \eta,\\
       &\mathop{\mathbb{E}}\limits_{\mathbf{G}\in\mathcal{G}}  \big(IoU[f(\mathbf{P}_t+(1-\mathbf{G}) \odot\boldsymbol{\Lambda} \odot \boldsymbol{\delta}_t, \mathbf{P}_0, \mathbf{B}_0), \mathbf{B}_t^*] \big) < \eta,\\
       &IoU[f(\mathbf{P}_t+\boldsymbol{\Lambda} \odot \boldsymbol{\delta}_t, \mathbf{P}_0, \mathbf{B}_0), \mathbf{B}_t^*] < \eta, \\
       &\boldsymbol{\Lambda} \in \{0, 1\}^{N_t \times 3}.
\end{aligned}
\end{equation}

The above problem is a mixed integer programming (MIP) problem. There are three challenges to optimizing it. The first one is the gradient calculation of the IoU operation, the second one is the non-differentiability of sparse mask $\boldsymbol{\Lambda}$ with the $l_0$-norm, and another one is the heavy computation of the expectation $\mathop{\mathbb{E}}$ under all $\mathbf{G}$. To this end, we propose a heuristic-based iteration update strategy.
Specifically, we replace the IoU function with the $\mathcal{H}$ proposed in Section \ref{sec:threat_models}. For the sparse attack term $\boldsymbol{\Lambda}$, we introduce a heuristic strategy to obtain the perturbation, which aims to drive the foreground points to deform subtly and keep the background points unchanged as much as possible. Finally, we take inspiration from the literature \cite{Liu2023Factorization}, and apply iteration updates to optimize the perturbation upon random factorized sub-vector at each time. 
The details are shown in Algorithm \ref{alg:solver}.

%% file: experiments.tex
\section{Experiments}\label{experiments}
In this section, we first present the LiDAR point cloud dataset used in this paper for evaluating adversarial attacks and the corresponding metrics to measure the success and imperceptibility of the attacks. Then, the relevant experimental details are described in detail. Next, we present the main experimental results, including white-box attack comparison results, black-box attack comparison results, and visualizations. Finally, the ablation study is reported to illustrate the effectiveness of the proposed transfer-based attack approach.

\subsection{Dataset}
This work uses two large-scale datasets from the point cloud tracking task: KITTI Tracking \cite{KITTI2013} and nuScenes \cite{nuscenes}. The former is a collection of 21 different scenarios using 64-beam LiDAR, containing a total of 808 tracking instances. The latter uses a 32-beam device to acquire 1,000 driving scenes, which are labeled with more than one million bounding boxes. For a fair comparison, we leveraged the same platform to retrain each victim model on the training set of both KITTI and nuScene and evaluated their robustness under adversarial attacks. To conduct well-rounded comparisons, we selected the most common target types from the dataset: pedestrians and cars.

\begin{figure*}[!ht]
     \centering
      \begin{minipage}[t]{0.24\linewidth}
        \centering
        \includegraphics[width=1\linewidth]{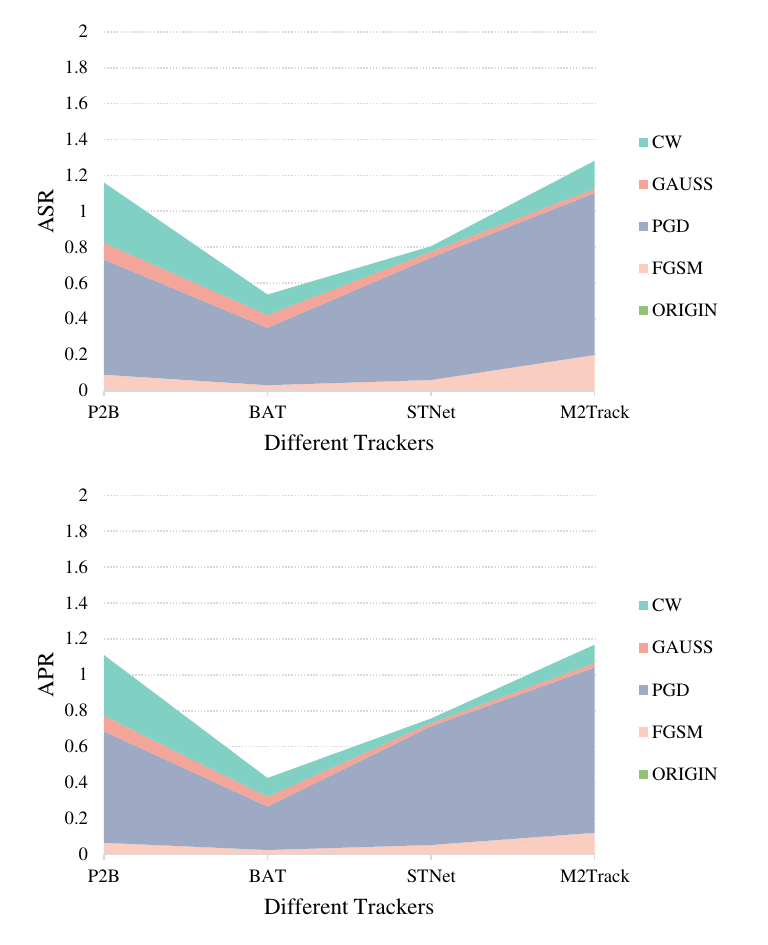}\\
        (a) Cars on KITTI
      \end{minipage}
      \begin{minipage}[t]{0.24\linewidth}
        \centering
        \includegraphics[width=1\linewidth]{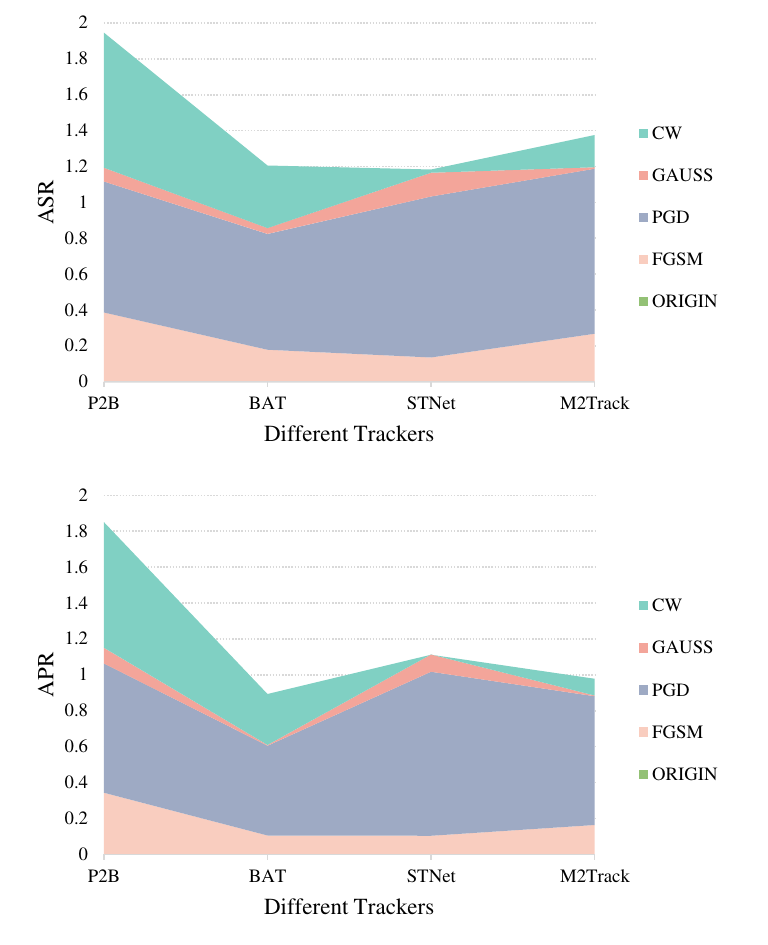}\\
        (b) Pedestrians on KITTI
      \end{minipage}
      \begin{minipage}[t]{0.24\linewidth}
        \centering
        \includegraphics[width=1\linewidth]{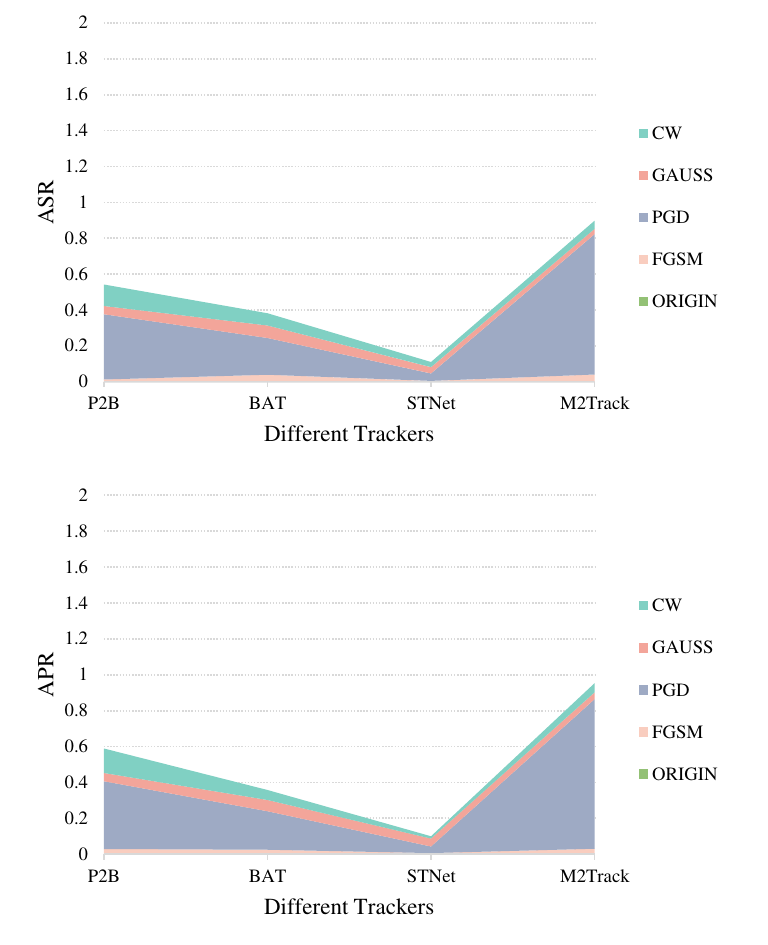}\\
        (c) Cars on nuScenes
      \end{minipage}
      \begin{minipage}[t]{0.24\linewidth}
        \centering
        \includegraphics[width=1\linewidth]{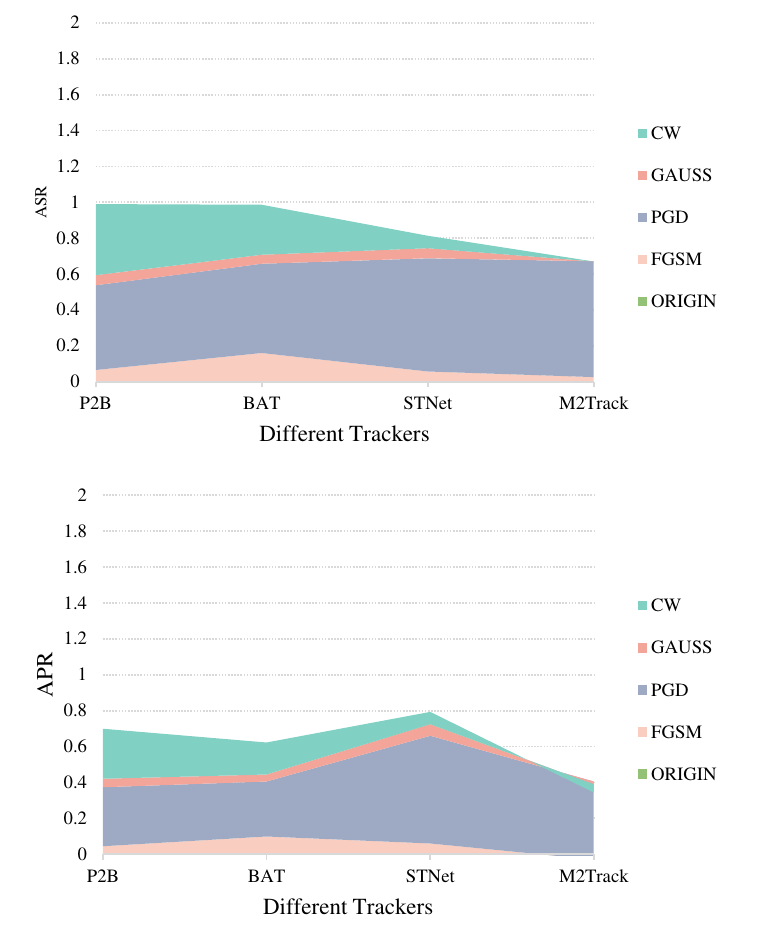}\\
        (d) Pedestrians on nuScenes
      \end{minipage}
    \caption{Plots of the attack success rate (ASR) and attack precision rate (APR). The cumulative areas of ASR and APR  are shown here. The larger the area, the better the results.}
    \label{fig:ASR_APR}
\end{figure*}

\renewcommand{\arraystretch}{1.5} 
\begin{table*}[!ht]
    \centering
    \belowrulesep=0pt
    \aboverulesep=0pt
    \caption{Experiments on cars and pedestrians of KITTI dataset. FGSM, PGD, and C\&W are white-box attack methods. GAUSS, TAN, and TAPG are black-box attack methods. The lower the SR and PR are, the more effective the attack is.}
    \label{tab:kitti}
    \begin{tabular}{c|ccccccccc}
    \toprule
        \multirow{3}{*}{Victim Models}&\multicolumn{8}{c}{Cars} \\
        \cmidrule(lr){2-9}
        ~ & \multicolumn{4}{c}{SR(\%)$\downarrow$} & \multicolumn{4}{c}{PR(\%)$\downarrow$} \\ 
        \cmidrule(lr){2-5} \cmidrule(lr){6-9}
        ~ & P2B\cite{P2B} & BAT\cite{BAT} & STNet\cite{stnet} & M2Track\cite{MMTrack} & P2B\cite{P2B} & BAT\cite{BAT} & STNet\cite{stnet} & M2Track\cite{MMTrack} \\ 
    \midrule
        ORIGIN & 56.82 & 63.29 & 67.33 & 67.54 & 71.10 & 75.93 & 78.76 & 81.18 \\
        FGSM\cite{FGSM} & 51.76 & 61.33 & 63.29 & 54.04 & 66.48 & 74.12 & 74.72 & 71.40 \\ 
        PGD\cite{PGD} & 20.35 & 43.06 & 21.36 & 6.56 & 27.02 & 57.51 & 26.62 & 6.41 \\ 
        C\&W\cite{CW} & 37.75 & 56.08 & 65.19 & 57.04 & 47.01 & 67.95 & 76.81 & 72.91 \\
        GAUSS & 51.51 & 58.71 & 65.23 & 65.96 & 64.9 & 71.85 & 77.29 & 79.17 \\ 
        TAN\cite{liu_transferable_2023} & 34.53 & 37.65 & 53.58 & 38.37 & 41.4 & 45.53 & 62.84 & 49.06 \\ 
        TAPG & 41.63 & 53.33 & 61.50 & 53.60 & 52.61 & 68.52 & 73.53 & 73.21 \\
    \midrule
        \multirow{3}{*}{Victim Models}&\multicolumn{8}{c}{Pedestrians} \\
        \cmidrule(lr){2-9}
        ~ & \multicolumn{4}{c}{SR(\%)$\downarrow$} & \multicolumn{4}{c}{PR(\%)$\downarrow$} \\ 
        \cmidrule(lr){2-5} \cmidrule(lr){6-9}
        ~ & P2B\cite{P2B} & BAT\cite{BAT} & STNet\cite{stnet} & M2Track\cite{MMTrack} & P2B\cite{P2B} & BAT\cite{BAT} & STNet\cite{stnet} & M2Track\cite{MMTrack} \\ 
    \midrule
        ORIGIN & 30.61 & 41.34 & 46.86 & 58.92 & 52.35 & 68.23 & 73.62 & 88.06 \\ 
        FGSM\cite{FGSM} & 18.76 & 33.94 & 40.49 & 43.13 & 34.41 & 61.12 & 66.02 & 73.68 \\ 
        PGD\cite{PGD} & 8.29 & 14.65 & 4.85 & 4.72 & 14.61 & 34.06 & 6.33 & 24.93 \\ 
        C\&W\cite{CW} & 7.58 & 26.9 & 46.01 & 48.35 & 15.68 & 48.86 & 73.73 & 79.89 \\ 
        GAUSS & 28.25 & 40.01 & 40.66 & 58.43 & 47.82 & 67.94 & 66.54 & 87.61 \\ 
        TAN\cite{liu_transferable_2023} & 7.78 & 11.62 & 13.82 & 18.32 & 16.7 & 27.15 & 32.22 & 47.57 \\ 
        TAPG & 12.84 & 33.16 & 37.99 & 52.49 & 25.83 & 57.71 & 61.81 & 83.42 \\    
    \bottomrule
    \end{tabular}
\end{table*}

\subsection{Metrics}
In previous studies, the performance of the tracker is mainly measured in terms of success rate (SR) and accuracy rate (PR). Specifically, the success rate is related to the IoU between the ground-truth box and the predicted box, which is calculated as the proportion of frames where the IoU exceeds a given threshold, and finally weighted and summed over different thresholds. Similarly, the accuracy is related to the error of the center coordinates between two boxes. To evaluate the robustness under adversarial attacks, we use the Attack Success Rate (ASR) and the Attack Precision Rate (APR), which essentially calculates the degraded values of SR and PR as a proportion of the original values:
\[
    ASR=\frac{SR-\widetilde{SR}}{SR}, 
    APR=\frac{PR-\widetilde{PR}}{PR}.
\]

In addition, to measure the imperceptibility of the adversarial samples, we utilize the Hausdorff distance (HD) and Chemfer distance (CD) to measure the discrepancy between the attacked sample and the original sample.
Let $\mathbf{P}_t$ and $\Tilde{\mathbf{P}}_t $ be original and attacked samples, the HD can be written as:
\[
    HD = \max\bigg\{
    \max_{\mathbf{p} \in \mathbf{P}_t} \{\min_{\Tilde{\mathbf{p}} \in \Tilde{\mathbf{P}}_t} \|\mathbf{p} - \Tilde{\mathbf{p}}\|_2 \}, 
    \max_{\Tilde{\mathbf{p}} \in \Tilde{\mathbf{P}}_t} \{\min_{\mathbf{p} \in \mathbf{P}_t} \|\mathbf{p} - \Tilde{\mathbf{p}}\|_2 \}
    \bigg\}.
\]
The CD can be written as:
\[
    CD = \frac{1}{|\mathbf{P}_t|}\sum_{\mathbf{p} \in \mathbf{P}_t} \min_{\Tilde{\mathbf{p}} \in \Tilde{\mathbf{P}}_t} \|\mathbf{p} - \Tilde{\mathbf{p}}\|_2 + \frac{1}{|\Tilde{\mathbf{P}}_t|}\sum_{\Tilde{\mathbf{p}} \in \Tilde{\mathbf{P}}_t} \min_{\mathbf{p} \in \mathbf{P}_t} \|\mathbf{p} - \Tilde{\mathbf{p}}\|_2.
\]
Notably, the HD aims to find the biggest gap between point clouds, which makes it capture the deformation in local geometric. The CD takes into consideration all points, being capable of measuring the variations of global geometric.

\subsection{Implementation Details}
The deep learning framework used in this work is Pytroch, which runs on the Ubuntu Linux system and is equipped with two 3090 NVIDIA GPUs and a Core i9 CPU.
For the pre-trained weights of the four victim models, we reproduce them on the above computing devices according to the official code released by the authors, and the parameter configurations are all set by default.
For the white-box adversarial attack, the center error $\sigma$ between two boxes in Eq. (3) is set to 0.3. Following the V2B \cite{V2B}, the $\beta$ and $\gamma$ of the focal loss in Eq. (4) are set to 2 and 4, respectively.
For the proposed black-box attack method, \ie, in Algorithm 1, the number of iterations $N_{iter}$ is set to 20, the parameter $\lambda$ in the surrogate loss is set to 0.5, and the optimization method is chosen to be Adam\cite{adam} with a learning rate $\xi$ of 0.01. 

\renewcommand{\arraystretch}{1.5} 
\begin{table*}[!ht]
    \centering
    \belowrulesep=0pt
    \aboverulesep=0pt
    \caption{Experiments on cars and pedestrians of nuScene dataset. FGSM, PGD, and C\&W are white-box attack methods. GAUSS, TAN, and TAPG are black-box attack methods. The lower the SR and PR are, the more effective the attack is.
 }
    \label{tab:nuscenes}
    \begin{tabular}{c|ccccccccc}
    \toprule
        \multirow{3}{*}{Victim Models}&\multicolumn{8}{c}{Cars} \\
        \cmidrule(lr){2-9}
        ~ & \multicolumn{4}{c}{SR(\%)$\downarrow$} & \multicolumn{4}{c}{PR(\%)$\downarrow$} \\ 
        \cmidrule(lr){2-5} \cmidrule(lr){6-9}
        ~ & P2B\cite{P2B} & BAT\cite{BAT} & STNet\cite{stnet} & M2Track\cite{MMTrack} & P2B\cite{P2B} & BAT\cite{BAT} & STNet\cite{stnet} & M2Track\cite{MMTrack} \\ 
    \midrule
        ORIGIN & 40.01 & 40.75 & 40.56 & 57.22 & 43.48 & 43.41 & 44.36 & 65.75 \\ 
        FGSM\cite{FGSM} & 39.5 & 39.14 & 40.33 & 54.84 & 42.21 & 42.3 & 44.07 & 63.72 \\ 
        PGD\cite{PGD} & 25.44 & 32.39 & 38.87 & 12.48 & 27.06 & 34.11 & 42.73 & 10.94 \\ 
        C\&W\cite{CW} & 35.24 & 37.98 & 39.39 & 54.64 & 37.54 & 41.01 & 43.77 & 62.35 \\ 
        GAUSS & 38.15 & 37.89 & 39.14 & 55.5 & 41.51 & 40.69 & 42.41 & 63.28 \\ 
        TAN\cite{liu_transferable_2023} & 35.37 & 35.24 & 36.71 & 51.45 & 37.91 & 37.75 & 40.13 & 60.75 \\ 
        TAPG & 36.12 & 37.41 & 39.57 & 53.59 & 38.75 & 40.56 & 43.62 & 63.10 \\ 
    \midrule
        \multirow{3}{*}{Victim Models}&\multicolumn{8}{c}{Pedestrians} \\
        \cmidrule(lr){2-9}
        ~ & \multicolumn{4}{c}{SR(\%)$\downarrow$} & \multicolumn{4}{c}{PR(\%)$\downarrow$} \\ 
        \cmidrule(lr){2-5} \cmidrule(lr){6-9}
        ~ & P2B\cite{P2B} & BAT\cite{BAT} & STNet\cite{stnet} & M2Track\cite{MMTrack} & P2B\cite{P2B} & BAT\cite{BAT} & STNet\cite{stnet} & M2Track\cite{MMTrack} \\ 
    \midrule
        ORIGIN & 24.96 & 27.24 & 28.89 & 34.75 & 46.44 & 49.82 & 55.48 & 60.35 \\
        FGSM\cite{FGSM} & 23.31 & 22.86 & 27.23 & 33.83 & 44.39 & 44.91 & 52.18 & 62.11 \\
        PGD\cite{PGD} & 13.18 & 13.7 & 10.67 & 12.34 & 31.15 & 34.51 & 22.16 & 34.14 \\ 
        C\&W\cite{CW} & 15.07 & 19.68 & 26.91 & 34.75 & 33.53 & 40.96 & 51.69 & 63.14 \\ 
        GAUSS & 23.56 & 25.86 & 27.25 & 34.78 & 44.25 & 47.9 & 51.89 & 61.26 \\ 
        TAN\cite{liu_transferable_2023} & 20.75 & 21.75 & 23.74 & 28.51 & 42.24 & 44.67 & 50.80 & 56.88 \\ 
        TAPG & 17.61 & 21.06 & 22.68 & 33.44 & 37.3 & 42.32 & 47.48 & 60.06 \\  
    \bottomrule
    \end{tabular}
\end{table*}

\begin{figure}[!t]
    \centering
    \includegraphics[width=1\linewidth]{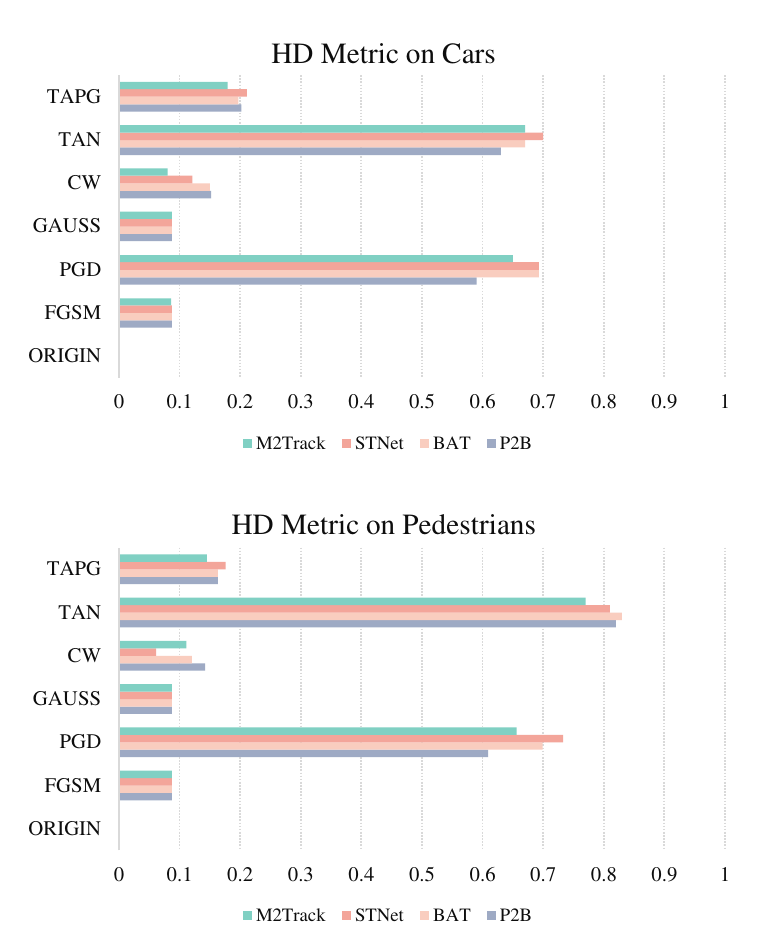}
    \caption{Hausdorff distance between origin and contaminated samples. Each 3D tracker (marked with different colors) is evaluated by six attack approaches. The first row shows the results for cars, and the second row shows the results for pedestrians.}
    \label{fig:HD}
\end{figure}
\begin{figure}[!t]
    \centering
    \includegraphics[width=1\linewidth]{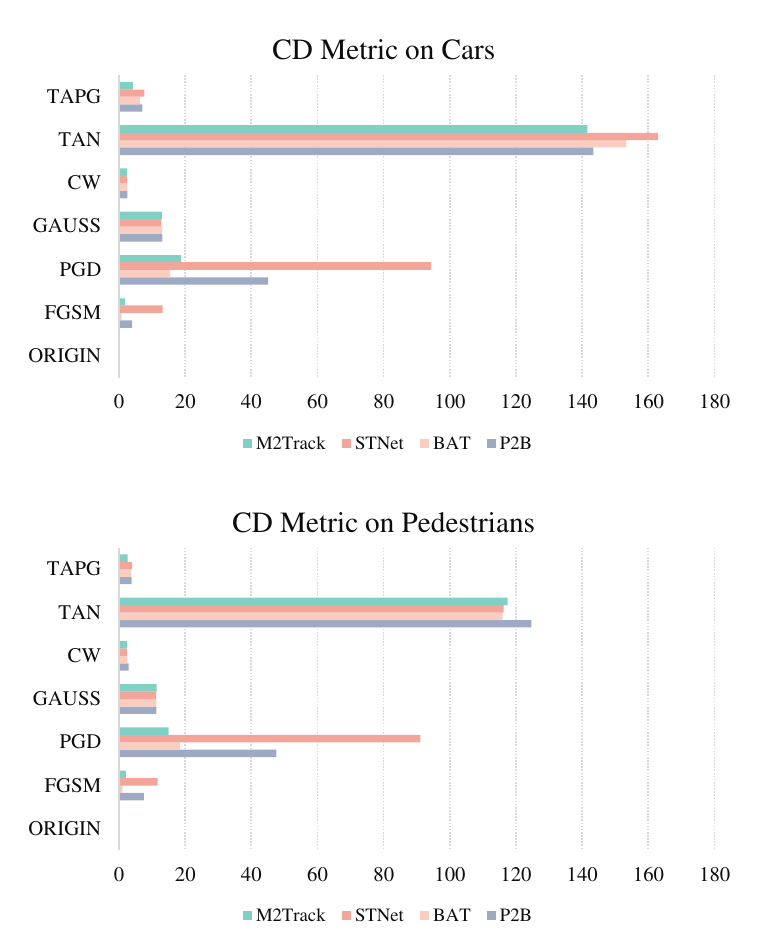}
    \caption{Chamfer distance between origin and contaminated samples. Each 3D tracker (marked with different colors) is evaluated by six attack approaches. The first row shows the results for cars, and the second row shows the results for pedestrians.}
    \label{fig:CD}
\end{figure}

\subsection{White-box Attack Results}\label{sec:exp_white}
Over the past few years, image classification has witnessed the continuous development of adversarial attacks. Among the representative methods of white-box attacks are FGSM with the help of gradient direction, PGD with iterative gradient direction update, and optimization-based C\&W. To evaluate the robustness of the current point cloud tracking models against the attacks, this paper starts from the designed loss function, \ie, Eq. (\ref{eq:attack_p2b}), (\ref{eq:attack_stnet}), (\ref{eq:attack_m2track}), and applies the above three attack methods to four point cloud trackers: P2B, BAT, STNet, and M2Track.

\subsubsection{Results on the KITTI Dataset}
In the first four rows of Table~\ref{tab:kitti}, we report the Success Rate (SR) and Precision Rate (PR) metrics of the origin performance of tracking models and the tracking performance after being attacked by FGSM, PGD, and C\&W, respectively. Overall, the current tracking models show a significant performance degradation when confronting white-box attacks on both pedestrian and car categories. In particular, PGD causes M2track's SR metric to decrease from 67.54\% to 6.56\% and PR metric to decrease from 81.18\% to 6.41\%. C\&W makes P2B collapse from 56.82\% to 37.75\% in SR metric and 71.10\% to 47.01\% in PR metric. This indicates that the current tracking model only focuses on improving accuracy and ignores the robustness of the model, which is a great security risk in real applications.

In addition, Fig. \ref{fig:ASR_APR} shows the ASR and APR metrics of the victim model under different attack methods. The figure can be interpreted in terms of both attack methods and tracking paradigms. On the one hand, from the perspective of evaluating different attack methods, we use different colors to indicate the attack methods, and the larger area occupied by a certain color indicates that its corresponding attack method is more effective. From Figure \ref{fig:ASR_APR}(a) and \ref{fig:ASR_APR}(b), it can be seen that the PGD method has the highest attack success rate on the KITTI, with C\&W in second place and FGSM in third place. On the other hand, from the perspective of evaluating the robustness of different tracking paradigms, we consider the cumulative area values from all white-box methods. The larger value of the accumulated ASR or APR indicates that the tracking model is more vulnerable. From Fig. \ref{fig:ASR_APR}(a) and \ref{fig:ASR_APR}(b), we can also find that the motion-based one-stream method, M2track, is the most sensitive than other two tracking paradigms in the car category; and all methods have a high cumulative APR and ASR, with vote-base two-stream paradigm (\ie, P2B) being the most sensitive.

\subsubsection{Results on the nuScenes Dataset}
Similarly, Table \ref{tab:nuscenes} gives the attack results on the nuScenes dataset. On the car category, compared to the original tracking results, the SR and PR of all victim models show a certain decrease after being attacked by FGSM, PGD, and CW. On the pedestrian category, all 3D trackers except M2track obtain different degrees of reduction in SR and PR metrics. Notice that the tracking models for the nuScenes are more robust than the KITTI. We believe this is mainly because the nuScenes data is more sparse than KITTI, which trains models that are resistant to the slight perturbations generated by the attack methods. In other words, the target distribution learned by the model in the sparse point cloud scenario is more inclusive, and it allows the LiDAR points to be shifted within a certain range. 

As can be seen in Fig. \ref{fig:ASR_APR}(c) and \ref{fig:ASR_APR}(d), PGD method also occupies the largest area on the nuScene and still maintains the highest attack effectiveness. From the perspective of tracking paradigms, the motion-based one-stream tracker has the largest cumulative ASR and APR, proving its vulnerability on the nuScene dataset. 
The BEV-based two-stream tracker, STNet, have a high cumulative APR when tacking contaminated pedestrians. Moreover, the vote-based two-stream trackers P2B and BAT also show strong sensitivity according to pedestrians' ASR.

\subsection{Black-box Attack Results}\label{sec:exp_black}
The above white-box attack approaches reflect the lack of robustness of current tracking models. However, since it is difficult to obtain exhaustive information about 3D trackers in practice, it would be more comprehensive to assess their robustness based on black-box attack methods. In view of this, we give a transfer-based black-box attack method in Section \ref{sec:target_aware}, which designs a target-aware strategy and random sub-vector decomposition to enhance the transferability. Note that TAN \cite{liu_transferable_2023} is the pioneer in performing adversarial attacks for point cloud tracking, which directly trained an auto-encoder to generate perturbations. Because its source code has not been published, we reproduce it based on the literature and compare it with our TAPG method proposed in this paper. In addition, to further compare the effectiveness of the proposed method, we also introduce an improved Gaussian-based FGSM, denoted as GAUSS, which randomly samples vector directions from a Gaussian distribution and substitutes it for gradient directions computed by white-box methods.

\subsubsection{Results on the KITTI Dataset}
Table \ref{tab:kitti} also reports the results of GAUSS, TAN, and the proposed TAPG method on KITTI. Overall, all three black-box attack approaches have obtained considerable performance relative to the ORIGIN. It is worth mentioning that our TAPG outperforms FGSM and C\&W even when compared with the white-box attack methods in terms of SR metrics. Note that these white-box methods can perform adversarial attacks with full information about the victim models, while our TAPG cannot. Specifically, when evaluating STNet on cars, it is found that FGSM and C\&W can only reduce their SR metrics from 67.33\% to 63.29\% and 56.08\%, whereas TAPG has obtained a greater reduction of 61.50\%. When evaluating STNet on pedestrians, TAPG also gains consistent performance, dropping its SR metric from the original performance of 46.86\% to 37.99\%, which is also superior to FGSM's 40.49\% and C\&W's 46.01\%.

\subsubsection{Results on the nuScenes Dataset}
Table \ref{tab:nuscenes} reports these results on nuScenes. Compared to the white-box method, the proposed TAPG achieves competitive results with C\&W and FGSM. In particular, when conducting attacks on M2Track, TAPG obtains more degradation of 53.59\% according to the SR metric in cars while C\&W and FGSM remain 54.64\% and 54.84\% respectively.

When compared to the black-box attack methods, our TAPG significantly outperforms the GAUSS on both the KITTI and nuScenes, demonstrating that the target-aware attack strategy in this paper is more effective than the stochastic gradient direction. Notably, our TAPG is slightly inferior to TAN. The reason lies in that the proposed TAPG is an online heuristic real-time attack, while TAN trains the perturbation generator offline and the large amount of data empowers it with adequate priority. Although TAN brings a higher attack success rate, it possesses two drawbacks. First, attacking different classes requires training the perturbation generator separately for each class, which brings about the consumption of computational resources for training, increases the deployment burden for testing, and reduces the attack flexibility. Second, the attack samples generated by TAN lack imperceptibility, \ie, they have a high Hausdorff distance and Chamfer distance relative to the original data in both local and global geometries. We will further discuss the comprehensive attribute of each attack method in the next section.

\begin{figure}[!t]
    \centering
    \includegraphics[width=1\linewidth]{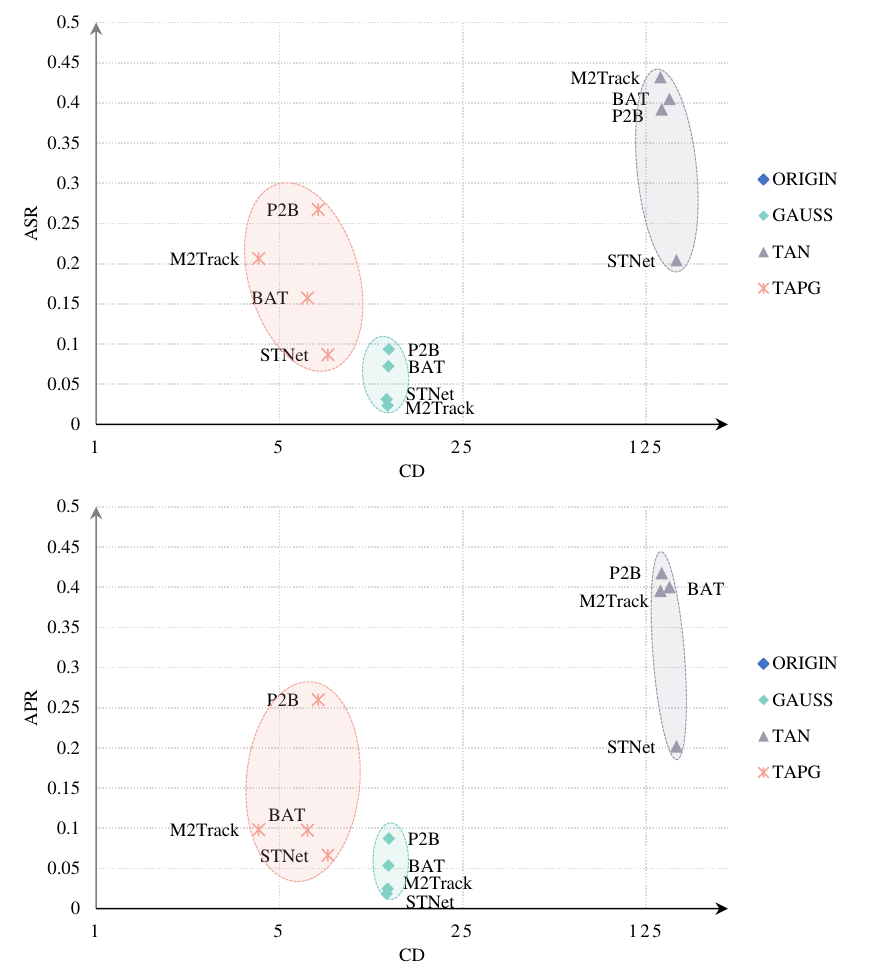}
    \caption{Plots of the CD and attack metrics on cars. The upper sub-figure is a scatter plot, with the X-axis being the CD values and the Y-axis being the ASR. The bottom sub-figure is plotted with the Y-axis representing the APR. Adversarial attack methods are employed in the context of diverse tracking paradigms.}
    \label{fig:bubble_car}
\end{figure}
\begin{figure}[!t]
    \centering
    \includegraphics[width=1\linewidth]{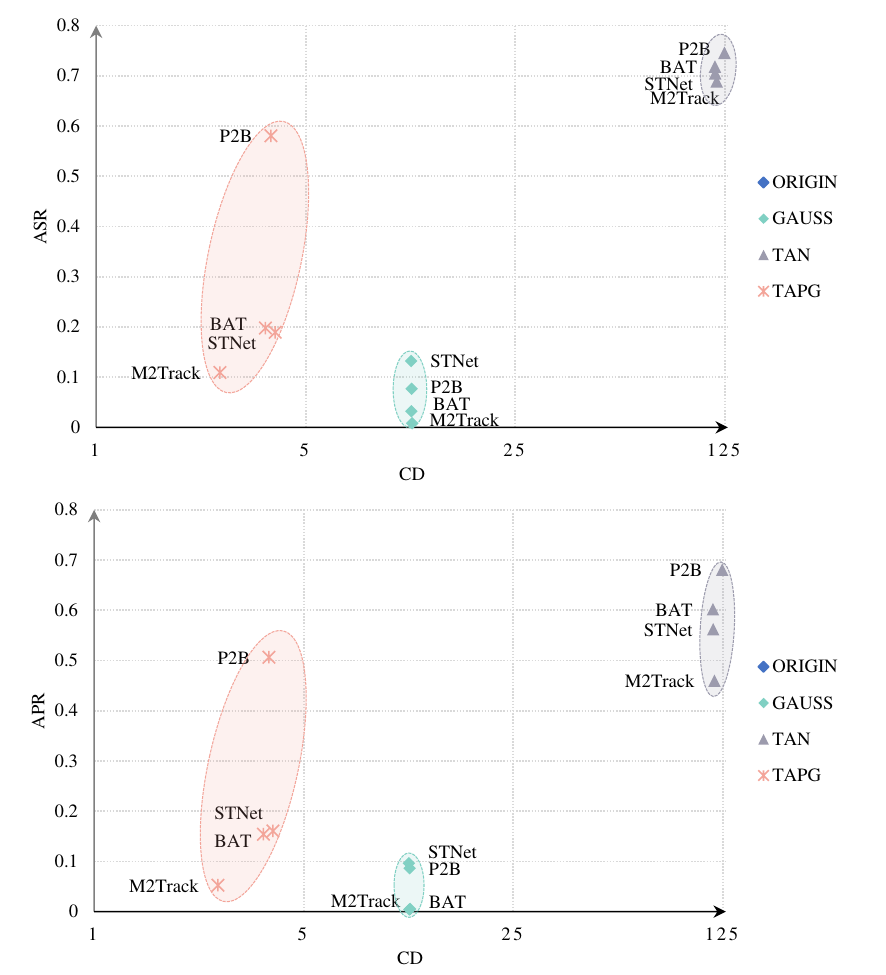}
    \caption{Plots of the CD and attack metrics on pedestrians. The top sub-figure is a scatter plot, with the X-axis being the CD values and the Y-axis being the ASR. The bottom sub-figure is plotted with the Y-axis representing the APR. Adversarial attack methods are employed in the context of diverse tracking paradigms.}
    \label{fig:bubble_ped}
\end{figure}

\begin{figure*}[ht]
    \centering
    \includegraphics[width=1\linewidth]{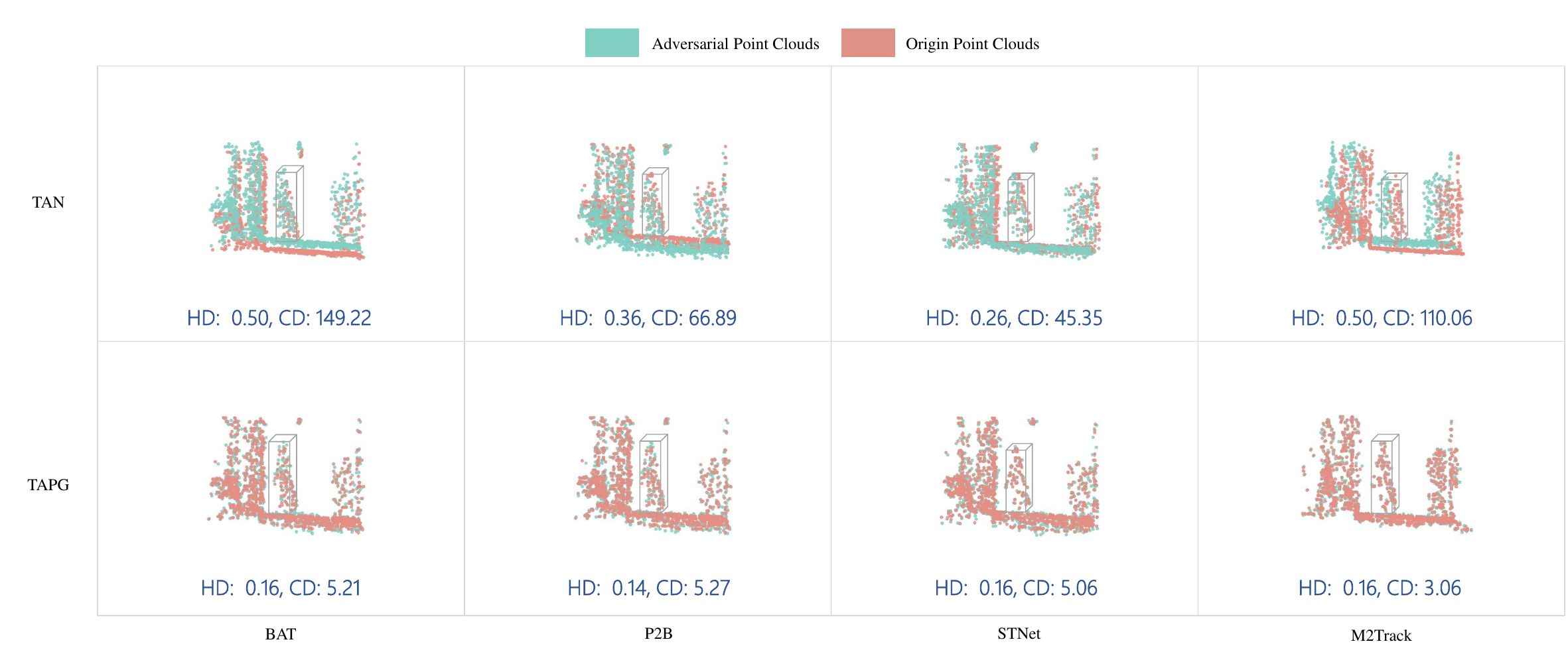}
    \caption{Visualization comparison between Hausdorff distance and Chamfer distance. The results of tracking pedestrian targets with various 3D trackers are presented in this figure. The green points are the disturbed adversarial point clouds and the red points are the original point clouds.}
    \label{fig:visual_HD_CD}
\end{figure*}

\renewcommand{\arraystretch}{1.5} 
\begin{table*}[!ht]
    \centering
    \caption{Ablation study of the proposed TAPG. The success rate (SR) and precision rate (PR) of four trackers are reported. The term ``sub-vector" is defined as a random sub-vector factorization. The term ``target-aware" is used to describe heuristic target-aware perturbation generation.}
    \label{tab:tapg_ablation}
    \belowrulesep=0pt
    \aboverulesep=0pt
    \begin{tabular}{cc|cccccccc}
    \toprule
        \multirow{2}{*}{Sub-vector} & \multirow{2}{*}{Target-aware} & \multicolumn{2}{c}{BAT\cite{BAT}} & \multicolumn{2}{c}{P2B\cite{P2B}} & \multicolumn{2}{c}{STNet\cite{stnet}} & \multicolumn{2}{c}{M2Track\cite{MMTrack}} \\
        \cmidrule(lr){3-4} \cmidrule(lr){5-6}\cmidrule(lr){7-8} \cmidrule(lr){9-10}
        ~ &~ & SR(\%) & PR(\%) & SR(\%) & PR(\%) &SR(\%) & PR(\%) & SR(\%) & PR(\%)\\
    \midrule
        ~          & ~          & 61.96 & 74.76 & 51.26 & 65.22 & 65.76 & 77.68 & 64.33 & 78.67 \\
        \checkmark & ~          & 59.81 & 72.81 & 45.06 & 57.80 & 65.50 & 76.94 & 63.55 & 78.25 \\
        ~          & \checkmark & 55.94 & 71.43 & 46.37 & 58.62 & 65.40 & 78.06 & 54.11 & 73.41 \\
        \checkmark & \checkmark & 53.33 & 68.52 & 41.63 & 52.61 & 61.50 & 73.53 & 53.60 & 73.21 \\
    \bottomrule
    \end{tabular}
\end{table*}

\begin{figure*}[ht]
    \centering
    \includegraphics[width=1\linewidth]{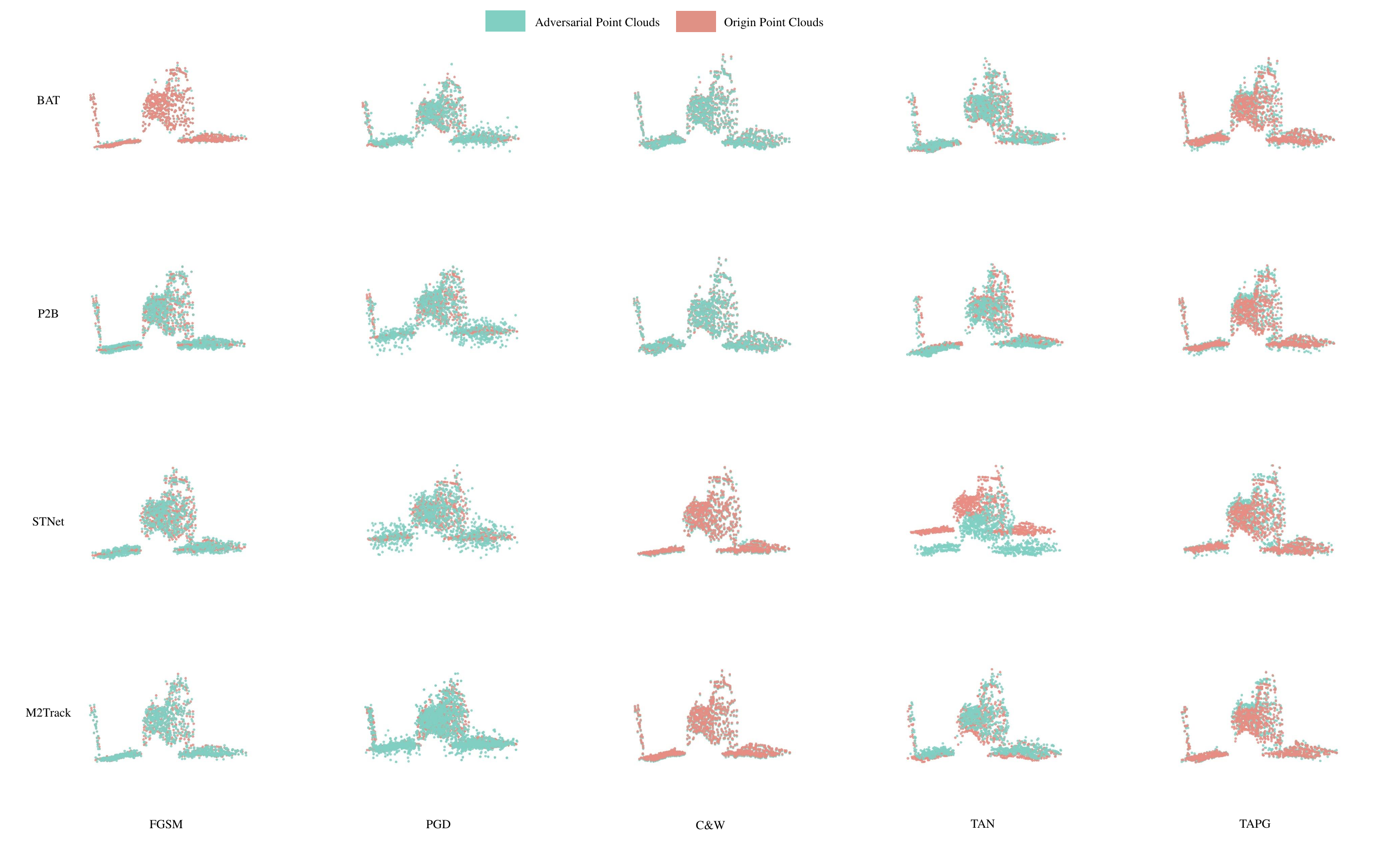}
    \caption{Visualization of adversarial samples generated by different attack methods. The targets represented by LiDAR point clouds are cars. The adversarial point clouds are plotted with green color, and the origin point clouds are plotted with red color.}
    \label{fig:visual_car}
\end{figure*}
\subsection{Study of Imperceptible Property}
In order to investigate their concealment ability, Fig. \ref{fig:HD} and \ref{fig:CD} plot the Hausdorff distance (HD) and the Chamfer distance (CD), respectively. The HD is used to identify the largest discrepancy between point clouds, thereby enabling the capture of deformation in local geometric properties. In contrast, the CD considers all points, allowing for the measurement of variations in global geometric characteristics. 
From these two figures, we summarize four findings for white-box and black-box attack methods. 
\begin{itemize}
    \item Firstly, among the white-box attack methods, FGSM method produces adversarial samples with the smallest HD and CD metrics, suggesting that it is the least perceptible; whereas the PGD, despite its high attack success rate in the Section \ref{sec:exp_white}, corresponds to a distance metric that is at least five times higher than that of the FGSM.
    \item  Second, among the black-box attack methods, TAN method obtains the maximum value on both HD and CD, indicating that it is extremely susceptible to perception despite its high attack rate (See the Section\ref{sec:exp_black}). This is inconsistent with our motivation introduced in the introduction. Unlike this, the proposed TAPG achieves a good balance between concealment and attack success.
    \item  Third, TAPG is higher than GAUSS in HD metrics and significantly lower than in CD metrics. This demonstrates that TAPG can control the interference of the adversarial samples better holistically. We deem that evaluating perturbations from a global geometric perspective (\ie, CD) is a better indicator of the degree of deformation than considering only the maximum difference (HD) produced by a single point.
    \item  Finally, according to the CD metric, TAPG can even achieve comparable results with some white-box attack methods (\eg, C\&W and FGSM).
\end{itemize}

The black-box attack methods are more compatible with practical application scenarios. In order to more intuitively see the coherence in terms of attack performance and imperceptibility performance, scatter plots of different black-box attack methods on the cars and pedestrians are given in Fig. \ref{fig:bubble_car} and Fig. \ref{fig:bubble_ped}, respectively. Following the principles of lower perceptibility and higher attack success, the ideal attack method should be near the top left corner of this figure. Overall, from these two plots, the proposed TAPG method attains a basic goal that it outperforms TAN and GAUSS methods in perceptibility and obtains competitive results in ASR and APR. Specifically, we have marked trackers attacked by TAPG as orange, which can be seen to have a more pronounced tendency to be closer to the top left relative to the other two methods.

\subsection{Visualization}
Experiments on quantitative analysis of different methods are reported in the above sections and some visualization results will be shown and analyzed below.

Fig. \ref{fig:visual_HD_CD} shows a visual comparison of the point cloud before and after the black-box attack, and gives the HD and CD values for each sample. From this, the following two questions may be of interest: 1) why the two metrics, HD and CD, are numerically different from each other; and 2) how the proposed TAPG achieves a significant decrease in the CD metric compared to the TAN method. To address the first question, the HD is a measure that identifies the point with the largest difference from two point sets (\ie, local geometry), whereas the CD calculates the sum of all paired points (\ie, global geometry). To address the second issue, we find that point clouds attacked by the TAN method tend to exhibit an overall shift, which leads to a larger perturbation at each point and thus an increase in the overall geometry. In addition, intuitively viewing the adversarial points (green) and the original points (red), the samples attacked by the TAN method largely have the green points overlaying the red points, whereas TAPG can better fit the distribution of the original point cloud. 

Furthermore, Fig. \ref{fig:visual_car} comprehensively compares the visualization results of the white-box and black-box attack methods applied to the four tracking models. From this figure, it can be seen that among the white-box attack methods, PGD method brings the largest deformation noise and is easily perceived. Among the black-box attack methods, the TAN method perceives the geometric variation more easily than our TAPG. Its attack characteristics are manifested as the translation transformation learned by the auto-encoder, which leads to higher CD and HD metrics. In the cross-comparison of black-box and white-box methods, the perceptibility of TAPG is comparable to that of FGSM and C\&W, and superior to that of PGD. 

\subsection{Ablation Study}
In this section, we perform ablation experiments on different strategies in the proposed TAPG to illustrate the effectiveness of each step in the attack algorithm. In addition, we replace TAPG with other white-box attacks to illustrate the necessity of special transfer-based attacks.

The proposed TAPG consists of two main strategies: a target-aware sparse attack strategy and an iterative update based on random sub-vector factorization. Table \ref{tab:tapg_ablation} gives the experimental results for the different strategies. The first row reports the results when neither strategy is used, from which we can see that the tracking performance remains high, \ie, the attack performance is the worst. When either ``sub-vector'' or ``target-aware'' is used, the attack capability is improved in both cases. Finally, when both strategies are adopted together, it can be observed that the attack performance obtains a significant improvement on the four typical trackers. Specifically, compared to the version where neither strategy is used, the SR metric of BAT drops from 61.96\% to 53.33\%, M2Track from 64.33\% to 53.60\%, and STNet from 65.76\% to 61.50\%. This speaks volumes about the effectiveness of each strategy.

To illustrate the necessity of the proposed black-box attack method, we compare it directly with the transfer-based attack using the white-box principle. Specifically, P2B serves as a surrogate model on which FGSM, PGD and CW are implemented. Then, we use the adversarial samples generated by this process to attack BAT, STNet and M2Track directly. Table \ref{tab:transfer} demonstrates the results of these transfer-based attacks. It can be seen that TAPG significantly outperforms FGSM and CW in terms of attack performance and obtains comparable results for its imperceptibility. When compared with PGD, the attack performance of TAPG is not as good as it, but its imperceptibility is significantly better than it. In general, TAPG achieves a better balance between attack capability and imperceptibility.

\renewcommand{\arraystretch}{1.5} 
\begin{table}[!ht]
    \centering
    \caption{Transferability validation experiments. The surrogate model is P2B and the adversarial samples are generated by different attack methods. we reported four metrics: the success rate (SR), precision rate (PR), Chamfer distance (CD), and Huasdorff distance (HD). The lower the four metrics are, the better the transferability is.}
    \label{tab:transfer}
    \belowrulesep=0pt
    \aboverulesep=0pt
    \begin{tabular}{cc|cccc}
    \toprule
        \makecell{Attack \\ Methods} & \makecell{Victim \\ Models} & SR(\%) $\downarrow$ & PR(\%) $\downarrow$ & CD $\downarrow$ & HD $\downarrow$ \\
    \midrule
        \multirow{3}{*}{FGSM} & BAT & 61.53 & 75.20 & 5.18 & 0.09 \\
        ~ & STNet & 66.37 & 77.86 & 4.73 & 0.09 \\
        ~ & M2Track & 65.94 & 79.35 & 5.29 & 0.09 \\ \hline
        \multirow{3}{*}{PGD} & BAT & 36.99 & 56.15 & 49.56 & 0.63 \\
        ~ & STNet & 38.65 & 48.18 & 38.56 & 0.62 \\
        ~ & M2Track & 34.72 & 48.92 & 55.01 & 0.64 \\ \hline
        \multirow{3}{*}{C\&W} & BAT & 62.80 & 76.28 & 2.58 & 0.67 \\
        ~ & STNet & 67.95 & 79.55 & 2.60 & 0.15 \\
        ~ & M2Track & 62.98 & 77.86 & 2.20 & 0.13 \\ \hline
        \multirow{3}{*}{GAUSS} & BAT & 58.71 & 71.85 & 13.03 & 0.087 \\
        ~ & STNet & 65.23 & 77.29 & 12.83 & 0.09 \\
        ~ & M2Track & 65.96 & 79.17 & 12.94 & 0.09 \\ \hline
        \multirow{3}{*}{TAPG} & BAT & 53.33 & 68.52 & 6.39 & 0.19 \\
        ~ & STNet & 61.50 & 73.53 & 7.66 & 0.21 \\
        ~ & M2Track & 53.60 & 73.21 & 4.16 & 0.18 \\
    \bottomrule
    \end{tabular}
\end{table}

%% file: conclution.tex
\section{Conclusion}
In this work, we investigate the robustness of current state-of-the-art point cloud tracking methods in the face of adversarial attacks. First, we provide a unified framework for point cloud tracking. Based on this, we design special loss functions to lift the white-box attack methods for the image domain to LiDAR point clouds. Experiments on KITTI and nuScene show that the tracking performance of current point cloud tracking methods degrades significantly when subjected to white-box attacks, among which the PGD method has the best attack effect, but the adversarial samples it generates are easily perceptible. In addition, considering the accessibility of network structure information in real-world attacks, we also consider transfer-based black-box attacks and propose a target-aware perturbation generation method called TAPG. Experiments show that the degradation of the tracking performance when encountering black-box attacks is also not negligible. And the proposed TAPG method achieves a better balance between attack performance and imperceptibility compared to existing TAN methods. More remarkably, TAPG is a black-box attack method that even outperforms FGSM and C\&W, which use the white-box attack principle. In future works, how to improve the robustness of the tracking model as well as how to design an attack method that is not easily perceptible and at the same time has a strong attack performance are worthwhile issues to investigate.